\documentclass[lettersize,journal]{IEEEtran}

\usepackage{amsmath,amsfonts}
\usepackage{algorithm}
\usepackage{algorithmic}
\usepackage{array}
\usepackage[caption=false,font=normalsize,labelfont=sf,textfont=sf]{subfig} 
\usepackage{textcomp}
\usepackage{stfloats}
\usepackage{url}        
\usepackage{verbatim}
\usepackage{graphicx}
\usepackage{cite}
\usepackage{float}
\usepackage{booktabs}
\usepackage{threeparttable}
\usepackage{multirow}
\usepackage{enumitem}
\usepackage{lscape}

\usepackage[table,xcdraw]{xcolor}
\usepackage{colortbl} 

\definecolor{LightGray}{gray}{0.90}
\newcommand{\hl}[1]{\cellcolor{LightGray}{#1}}

\makeatletter
\renewcommand\p@subsubsection{} 
\makeatother
\usepackage[colorlinks=true,linkcolor=blue,citecolor=blue,urlcolor=black]{hyperref}
\usepackage{cleveref}

\begin{document}

\title{VFM-Guided Semi-Supervised Detection Transformer under Source-Free Constraints for Remote Sensing Object Detection}

\author{Jianhong Han,~\IEEEmembership{Student Member,~IEEE,} 
Yupei Wang$^{*}$,~\IEEEmembership{Member,~IEEE,} Liang Chen, ~\IEEEmembership{Member,~IEEE}

\IEEEcompsocitemizethanks{
\IEEEcompsocthanksitem This work was supported by National Natural Science Foundation of China under Grant 62301046, National Key Laboratory for Space-Born Intelligent Information Processing under Grant TJ-01-22-01.
\textit{
(Corresponding author: Yupei Wang.)}
\IEEEcompsocthanksitem J. Han, Y. Wang and L. Chen are with the School of Information and Electronics, Beijing Institute of Technology, Beijing 100081, China, also with the Beijing Institute of Technology Chongqing Innovation Center, Chongqing, 401135, China, and also with the National Key Laboratory for Space-Born Intelligent Information Processing, Beijing 100081, China.  (e-mail: hanjianhong1996@163.com;wangyupei2019@outlook.com; chenl@bit.edu.cn).
}
}

\maketitle

\begin{abstract}
Unsupervised domain adaptation methods have been widely explored to bridge domain gaps. However, in real-world remote-sensing scenarios, privacy and transmission constraints often preclude access to source domain data, which limits their practical applicability. Recently, Source-Free Object Detection (SFOD) has emerged as a promising alternative, aiming at cross-domain adaptation without relying on source data, primarily through a self-training paradigm. Despite its potential, SFOD frequently suffers from training collapse caused by noisy pseudo-labels, especially in remote sensing imagery with dense objects and complex backgrounds. Considering that limited target domain annotations are often feasible in practice, we propose a Vision foundation model-Guided DEtection TRansformer (VG-DETR), built upon a semi-supervised framework for remote sensing object detection under source-free constraints. VG-DETR integrates a Vision Foundation Model (VFM) into the online training pipeline in a \emph{free lunch} manner, leveraging a small amount of labeled target data to mitigate pseudo-label noise while improving the detector’s feature-extraction capability. Specifically, we introduce a VFM-guided pseudo-label mining strategy that leverages the VFM’s semantic priors to further assess the reliability of the generated pseudo-labels. By recovering potentially correct predictions from low-confidence outputs, our strategy improves pseudo-label quality and quantity. In addition, a dual-level VFM-guided alignment method is proposed, which aligns detector features with VFM embeddings at both the instance and image levels. Through contrastive learning among fine-grained prototypes and similarity matching between feature maps, this dual-level alignment further enhances the robustness of feature representations against domain gaps. Extensive experiments demonstrate that VG-DETR achieves superior performance in source-free remote sensing detection tasks. Code is released at \href{https://github.com/h751410234/VG-DETR}{https://github.com/h751410234/VG-DETR}.
\end{abstract}


\vspace{1cm}

\section{Introduction}
\IEEEPARstart{D}{eep} learning-based object detection in remote sensing imagery has been widely applied, achieving outstanding performance in recent years. However, such methods are constrained by data-driven supervised learning frameworks that require training and testing data to have the same distribution, which leads to performance degradation when a domain gap exists. To address this issue, Unsupervised Domain Adaptive (UDA) object detection\cite{Chen_Li_Sakaridis_Dai_Van_Gool_2018,Huang_Lu_Lin_Xie_Lin,Xu_Sun_Diao_Zhao_Fu_Wang_2022} has received significant attention, focusing on transferring knowledge from source domains to target domains by using labeled source domain data and unlabeled target domain images. In practical remote sensing scenarios, due to high data transmission costs or data privacy concerns, source domain data is often inaccessible, hindering the application of UDA methods.

To address the above shortcomings, Source-Free Object Detection (SFOD) methods\cite{li2021free,zhang2023multi,liu2024source} have recently emerged as a promising research direction. These methods adapt a source-trained model to target domain data by utilizing only unlabeled target domain images, thus better aligning with practical deployment requirements. Since source domain data is completely inaccessible, commonly used style transfer\cite{huang2017arbitrary,Zhao_Zhong_Zhao_Sebe_Lee_2022} and feature alignment techniques\cite{Chen_Li_Sakaridis_Dai_Van_Gool_2018,10335732,Xu_Sun_Yang_Miao_Yang_Research,yang2025fsda} are generally infeasible. Instead, self-training paradigms based on the mean teacher framework\cite{Liu_Ma_He_Kuo_Chen_Zhang_Wu_Kira_Vajda_2021} have become the dominant approach. These paradigms consist of teacher and student models with identical structures, where the teacher model generates pseudo-labels to guide the student model in supervised training. Subsequently, the teacher model's weights are updated through the Exponential Moving Average (EMA)\cite{Marsella_Gratch_2009} of the student model. Ultimately, this iterative training process enables the adaptation of the source-trained model to target domain data.

Despite notable performance gains, self-training frequently suffers from training collapse induced by noisy pseudo-labels, especially in cross-domain remote-sensing detection. As shown in subsection \ref{Training_Stability} of our experiments, the optimization curves of source-free training paradigms reveal that collapse is difficult to avoid. Although several recent methods\cite{liu2023periodically,khanh2024dynamic} aim to stabilize self-training for source-free object detection, they remain vulnerable in remote-sensing scenarios, where complex backgrounds and dense object distributions exacerbate pseudo-label errors compared with natural scenes. Prevailing evaluation protocols often rely on early stopping or on reporting peak accuracy on labeled test sets, which can obscure the severity of training collapse and are not well aligned with practical applications. Where labels are available, conventional supervised training is the more appropriate choice and offers a stronger performance.

To address the aforementioned problem, we found that employing a semi-supervised framework for SFOD tasks can effectively guide the learning of unlabeled data and lead to a stable training process under limited labeled supervision. Since a small amount of target domain data is permitted to be labeled in practical applications, leveraging this accurate information not only mitigates learning errors introduced by pseudo-labels but also significantly improves detection performance. To further unlock the detector’s potential, we consider leveraging a Vision Foundation Model (VFM) to guide its learning. This model combines large-scale architectures with self-supervised pretraining on massive datasets and has already demonstrated strong guidance capabilities across a wide range of computer vision tasks \cite{oquab2023dinov2,kirillov2023segment,liu2024grounding,10504785}. Rather than simply using foundation models as backbone, we utilize them as a reference to enhance the quality of pseudo-labels generation and enable more robust feature extraction for the detector. In addition, we incorporate the VFM into the online training pipeline in a \emph{free lunch} manner, thereby avoiding significant additional computational cost.

In this paper, we propose a Vision foundation model-Guided DEtection TRansformer (VG-DETR) built upon a semi-supervised framework under source-free constraints for remote sensing object detection. Our method integrates a VFM to effectively exploit a limited amount of labeled data, refine the quality of generated pseudo-labels, and enhance the detector’s feature representations, thereby achieving robust performance while avoiding training collapse.

To maintain high-quality and high-quantity pseudo-labels generation, we propose a \textbf{VFM-guided Pseudo-label Mining (VPM)} strategy. Unlike existing approaches based on fixed or dynamic thresholds, which often introduce excessive noise or overlook potentially correct predictions, our strategy incorporates an external semantic prior by leveraging robust features extracted from the VFM to identify likely correct predictions among low-confidence outputs. Specifically, during the offline stage, we extract feature maps from target domain images using a VFM. Then, the features with partial annotations are selected and clustered via K-means \cite{macqueen1967some} to derive robust class-wise reference prototypes and background prototypes. During training, following a similar procedure based on pseudo-labels, we extract instance-level features from unlabeled samples and compare them to the referenced prototypes via cosine similarity to estimate the reliability of the corresponding pseudo-labels from a semantic perspective. Finally, we integrate the auxiliary reliability with the detector’s confidence scores to identify low-confidence yet potentially correct pseudo-labels, enabling adaptive pseudo-label mining.

In addition, we propose a \textbf{Dual-level VFM-guided  Alignment (DVA)} approach that uses VFM features as semantic anchors to enhance the robustness of the detector's learned representations. Specifically, we first aggregate the detector’s object queries into multiple fine-grained prototypes via a Sinkhorn-based soft clustering\cite{cuturi2013sinkhorn} mechanism. These fine-grained prototypes are subsequently guided by the previously constructed class-wise reference prototypes from the VFM through contrastive learning, enabling the object queries to capture more generalizable instance-level feature representations within the decoder. At the image level, we adopt using the feature space of the VFM as a proxy for domain alignment. By independently computing the similarity between the detector’s backbone feature maps and those extracted from a frozen VFM, we enhance the consistency of their feature representations, thereby promoting robust learning and enhancing semantic clarity in foreground regions.

The main contributions of this paper are as follows:

\begin{enumerate}[label=\arabic{*})]
    \item We conduct a study of self-training paradigm under source-free constraints for remote-sensing object detection, revealing pronounced failure applications arising from training collapse. We then propose VG-DETR, a semi-supervised detection transformer designed for remote-sensing imagery, which demonstrates stable training and delivers superior performance.
    
    \item A VPM strategy is introduced, which innovatively incorporates external semantic priors extracted from a VFM to provide auxiliary evaluation of the generated pseudo-labels, thereby reducing reliance on the detector’s own predictions. This strategy enables the adaptive selection of potentially correct low-confidence predictions, ensuring that the generated pseudo-labels are of both high quality and sufficient quantity.
    
    \item We further propose a DVA method that aligns detector representations with VFM embeddings at both the instance and image levels. By softly clustering object queries into fine-grained prototypes and contrastively pulling them toward VFM prototypes, while simultaneously introducing a global alignment term that matches backbone feature maps to their VFM counterparts, this dual alignment strengthens the robustness of the detector’s representations against domain gaps.
\end{enumerate}

We conducted comprehensive experiments demonstrating that VG-DETR delivers superior performance and generalization compared with the current state of the art. Because source-free training can collapse, we adopt a semi-supervised framework that mitigates this issue by injecting a small amount of labeled data (1\%, 5\%, 10\%) across multiple remote-sensing adaptation scenarios. With only 5\% labeled data, VG-DETR achieves 77.5\% mAP on the classic cross-satellite adaptation task (xView\cite{Lam_Kuzma_McGee_Dooley_Laielli_Klaric_Bulatov_McCord_2018} → DOTA\cite{Xia_Bai_Ding_Zhu_Belongie_Luo_Datcu_Pelillo_Zhang_2018}), surpassing all competing methods under every setting. On the synthetic-to-real benchmark (SRSD → DIOR\cite{Li_Wan_Cheng_Meng_Han_2020}), our approach boosts mAP by more than 65.9\%, confirming its effectiveness. In the cross-modal adaptation setting (HRRSD\cite{zhang2019hierarchical} → SSDD\cite{Li_Qu_Shao_2017}), VG-DETR further demonstrates its generalization ability, attaining 70.6\% mAP.

\section{Related Work}

\subsection{Cross-Domain Object Detection in Remote Sensing Images}

Mainstream cross-domain object detection research typically follows an UDA paradigm, which narrows the gap between the source and target domains by jointly leveraging labeled source data and unlabeled target images. Chen et al. \cite{Chen_Li_Sakaridis_Dai_Van_Gool_2018} were the first to implement UDA on Faster R-CNN\cite{Ren_He_Girshick_Sun_2017}, applying adversarial training separately to the backbone and the region-proposal network to enable the detector to learn domain-invariant features, thereby laying the foundation for subsequent studies. With the success of DETR\cite{carion2020end}, researchers quickly extended it to cross-domain tasks, spawning a series of follow-up works. Wang et al. \cite{Wang_Cao_Zhang_He_Zha_Wen_Tao_2021} inserted a learnable domain query into the encoder, enabling the attention mechanism to adaptively aggregate domain-specific features, which were subsequently aligned through adversarial training. Zhang et al. \cite{zhang2023detr} proposed the CNN–Transformer blender, which fuses CNN and Transformer features to improve feature alignment in both the backbone and the encoder. 

Research on cross-domain object detection in remote‐sensing imagery has progressed relatively slowly. Xu et al.\cite{Xu_Sun_Diao_Zhao_Fu_Wang_2022} proposed the first UDA approach built on the one-stage detector RetinaNet\cite{Lin_Goyal_Girshick_He_Doll} and introduced an adaptive feature alignment strategy to address the challenge of accurately aligning sparse foreground objects. Zhu et al.\cite{zhu2023dualda} incorporated a mean-teacher framework into remote sensing adaptation, using dual detection heads to suppress biased information and refine pseudo-labels. Han et al.\cite{10474037} introduced the first DETR-based detector for UDA object detection in remote sensing, performing feature alignment in the frequency domain and leveraging learnable filters to adaptively select domain-invariant features.

Nevertheless, all of the above methods presuppose access to source-domain data. In practical remote sensing applications, high data-transfer costs and privacy restrictions often make such access infeasible. Consequently, this study investigates source-free cross-domain object detection, in which a pretrained detector is adapted to the target domain using only target-domain data.

\subsection{Source-Free Object Detection}

To address scenarios where source-domain data are unavailable, SFOD has recently garnered considerable attention. Most existing approaches are built on the mean teacher framework, whose chief aim is to improve the quality of generated pseudo-labels while preserving training stability. Li et al.\cite{li2021free} were the first to propose a source-free object detector and to develop a self-entropy based approach for setting an appropriate confidence threshold. Zhang et al.\cite{zhang2023multi} constructed multiple prototypes for instance features within a Faster R-CNN detector and leveraged the consistency between teacher- and student-prototype distributions to correct the generated pseudo-labels. Deng et al.\cite{deng2024balanced} leveraged prediction uncertainty to assess sample difficulty and adopted a curriculum-style training schedule to mitigate distribution variance across the training data. To address training instability, Liu et al.\cite{liu2023periodically} proposed a dual-teacher training strategy in which a dynamic and a static teacher are periodically exchanged to stabilize self-training. Trinh et al.\cite{khanh2024dynamic} devised a controlled-update scheme that compares current prediction uncertainties with their historical counterparts to determine appropriate timing for parameter updates, thereby minimizing the accumulation of learning errors.

Research on SFOD in the field of remote sensing remains limited. Liu et al.\cite{liu2024source} developed their method using the Faster R-CNN and introduced a perturbation module to augment feature styles by simulating pixel-level variations in color and noise commonly observed in remote sensing domains. In addition, similar to the methods described above, they considered using robust prototypes to correct noisy pseudo-labels. However, we observe that noisy pseudo-labels frequently trigger training collapse, especially in cross-domain remote-sensing object detection. Considering that a small amount of target-domain data can often be annotated in real-world scenarios, our method adopts a semi-supervised framework that leverages limited labeled data to supervise the learning of unlabeled images, thereby enabling a stable training process. Furthermore, we explore the effective integration of vision foundation models to further guide the learning process of the model, aiming to enhance the detector’s performance.

\subsection{Pseudo-Label Generation in Object Detection}

Generating pseudo-labels for unlabeled images is central to semi-supervised learning, as it enables unlabeled data to contribute to supervised training. Chen et al.\cite{chen2022dense} proposed an aggregated teacher that enhances the teacher model’s capacity by performing parameter aggregation across time and recurrent aggregation across layers, ultimately generating more accurate pseudo-labels. Zhou et al.\cite{zhou2022dense} introduced a dense pseudo-labeling approach that removes the traditional NMS\cite{neubeck2006efficient} post-processing, thereby preserving richer instance-level information. Xu et al.\cite{xu2021end} evaluated pseudo-label reliability based on confidence scores, applying a weighted mechanism to reduce the impact of low-confidence predictions during training. Han et al.\cite{10474037} assessed the model’s current learning status to adaptively determine suitable threshold values for pseudo-label selection. Fang et al.\cite{fang2024dual} incorporated two teacher models with different weights and architectures, merging their pseudo-labels to mitigate errors arising from architectural bias. Zhang et al.\cite{10969845} proposed a dynamic multiview learning strategy to address the inherent trade-off between pseudo-label quality and quantity, by separating predictions into multiple hierarchies for more fine-grained filtering.

Despite efforts to improve pseudo-label quality, most existing methods confine their evaluation to the detector itself, relying solely on internal heuristics to judge pseudo-label reliability. Moreover, existing semi-supervised DETR-based detectors\cite{wang2022omni,zhang2023semi,shehzadi2024sparse} still adopt fixed thresholds, largely overlooking the potential of pseudo-label mining. Recently, VFMs have demonstrated superior feature granularity and generalization capabilities. Motivated by this, we leverage VFM-extracted features to externally evaluate the reliability of pseudo-labels. By recovering potentially correct predictions from low-confidence outputs, we improve both the quality and quantity of the generated pseudo-labels. To the best of our knowledge, this is the first work to leverage a VFM to improve pseudo-label generation.

\subsection{VFM in object detection}

Recently, research on applying VFMs to various downstream tasks has attracted considerable attention due to the strong capabilities of the extracted features in both fine-grained representation and generalization. For object detection, Zhang et al.\cite{zhang2023detect} integrated a VFM into few-shot detection tasks by leveraging feature prototypes derived from its extracted components its extracted components, enabling more robust recognition of novel classes. Building on this, Fu et al.\cite{fu2024cross} extended the approach to cross-domain detection by transforming the extracted prototypes into learnable representations to better handle the challenges posed by domain gaps. Lavoie et al.\cite{lavoie2025large} leveraged a VFM to build a new labeller for unsupervised domain adaptation tasks, which was trained using source-domain annotations and produced more accurate pseudo-labels than those generated by a vanilla detector.

In contrast to the above approaches that simply treat the frozen foundation model as a backbone, which in turn constrains the detector’s architecture and limits its applicability in scenarios such as resource-constrained environments or source-free settings. Fu et al.\cite{fu2024frozen} explored using the foundation model as a plug-and-play module to guide supervised learning in the detector by leveraging its class token, which provides an in-depth understanding of complex scenes. Bi et al.\cite{bi2025good} observed that feature styles extracted from CLIP\cite{radford2021learning} exhibit stronger generalization capabilities, and thus incorporated them into the detector backbone to tackle domain generalization tasks. These methods still require the insertion of more or less carefully designed modules into the detection pipeline, often accompanied by additional inference overhead. In this paper, we explore a \emph{free lunch} style of guidance in the online training process of the detector by improving the quality of generated pseudo-labels and enhancing the detector’s feature extraction capabilities, thereby effectively boosting detection performance.

\section{Method}

\subsection{Problem Statement}
We first provide a brief problem statement that precisely defines the SFOD task. This task involves two distinct domains, \begin{math} D = \{D_{s} , D_{t}\} \end{math}, where \begin{math} D_{s} \end{math} denotes the source domain containing both images and annotations, and \begin{math} D_{t} \end{math} refers to the target domain, which provides only images. Conventional unsupervised domain adaptation methods enhance model performance by leveraging data from both \begin{math} D_{s} \end{math} and \begin{math} D_{t} \end{math}. In contrast, the SFOD task aims to adapt a source-trained model (trained on source domain data \begin{math} D_{s} \end{math}) to the target domain, using only \begin{math} D_{t} \end{math} without utilizing data from \begin{math} D_{s} \end{math}.

Building upon the SFOD setting, we introduce a semi-supervised extension in which a limited subset of \begin{math} D_{t} \end{math} samples (typically 1\%, 5\%, or 10\%) are allowed to have annotations, and this assumption is often considered practical in real-world applications. Under this semi-supervised source-free setting, our proposed method not only avoids training collapse but also achieves detection performance comparable to fully supervised counterparts.

\begin{figure*}[htbp]
\centering
\includegraphics[width=18cm]{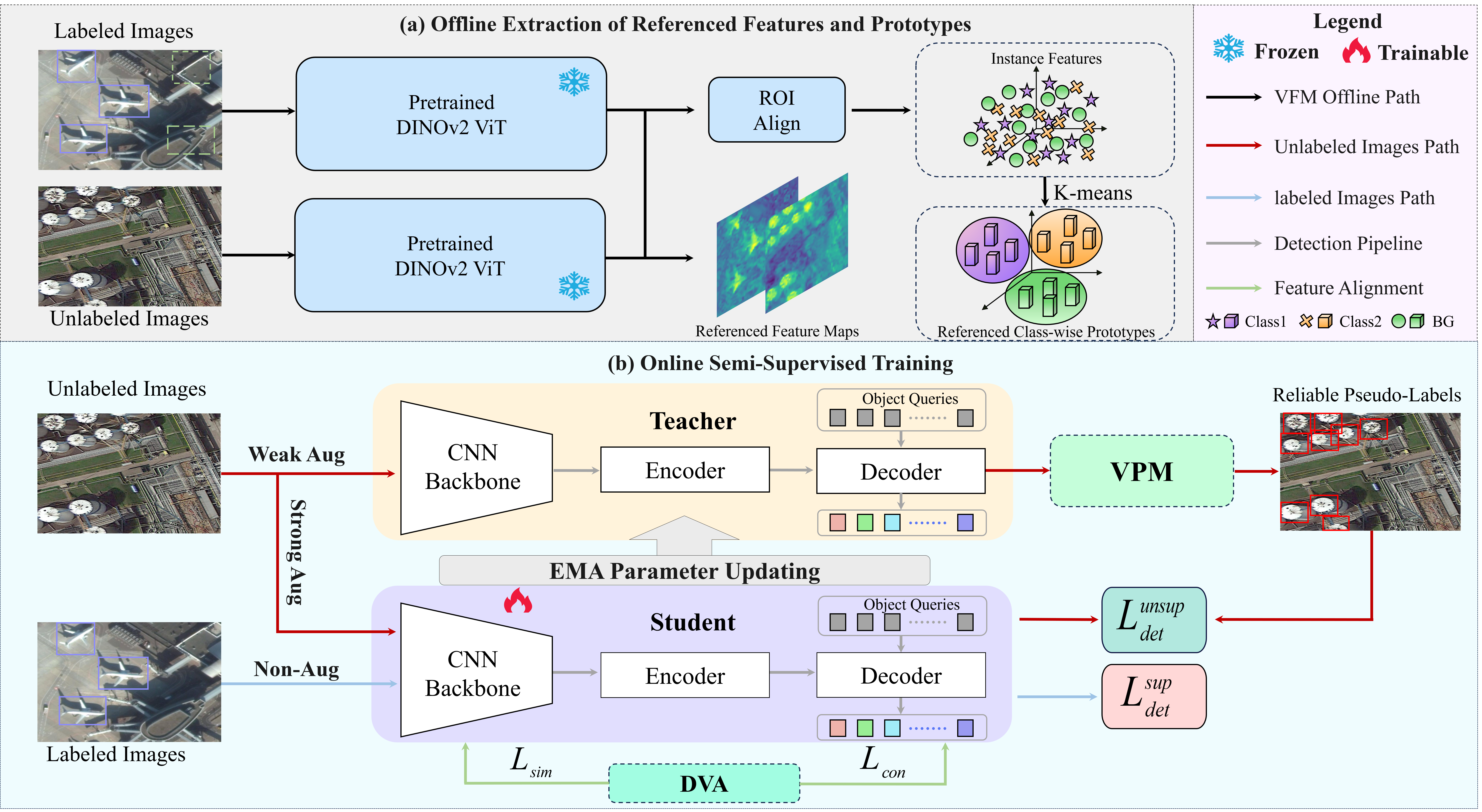}
\caption{
The overall VG-DETR pipeline operates in two stages. In the offline stage, reference feature maps and class-wise prototypes are extracted from the VFM using target-domain images. The online semi-supervised training stage employs two networks: a student detector and its temporally-ensembled counterpart, referred to as the teacher model. The teacher generates pseudo-labels for unlabeled target-domain images, enabling the student to learn from them. The proposed VPM strategy leverages the extracted prototypes to adaptively mine potentially correct predictions from low-confidence outputs. In addition, the designed DVA module jointly exploits the reference feature maps and prototypes to align detector representations with VFM embeddings at both the image and instance levels, thereby substantially improving the detector’s robustness in cross-domain scenarios.
}
\label{fig_1}
\end{figure*}

\subsection{Overview}
The self-training paradigm is a mainstream approach in SFOD tasks, leveraging pseudo-labels generated from target domain images to optimize a source-trained model. Our VG-DETR integrates the base detector with a mean-teacher self-training framework and further incorporates an additional supervised training branch to leverage the labeled samples under a semi-supervised setting. The overall framework is illustrated in Fig. \ref{fig_1} (b). Specifically, this framework employs two models with identical architectures: a teacher model and a student model, both initialized with source-trained weights. For the self-training branch on unlabeled data, the teacher model generates pseudo-labels from weakly augmented images, while the student model performs predictions on strongly augmented images. For the additional supervised training branch handling the limited labeled samples, we follow the vanilla DINO\cite{zhang2022dino} design. Then, the student model computes two types of loss, thereby facilitating adaptation to the target domain. The specific loss functions are defined as follows:
\begin{equation}
\begin{aligned}
\mathcal{L}_{\text {det }} &= \mathcal{L}_{\text {det }}^{\text {sup }} + \mathcal{L}_{\text {det }}^{\text {unsup }}  \\
  &= \mathcal{L}_{\text {det }}\left(P_{s}^{\text {sup}}, Y^{\text {sup}}\right) + \mathcal{L}_{\text {det }}\left(P_{s}^{\text {unsup}}, P_{t}^{\text {unsup}}\right),\\
\end{aligned}
\end{equation}
where \begin{math} \mathcal{L}_{det}^{\text {sup }} \end{math}  denotes the supervised loss corresponding to the additional supervised training branch, and \begin{math} \mathcal{L}_{det}^{\text {unsup }} \end{math} represents the unsupervised loss corresponding to the self-training branch. Both losses contain classification and regression terms. \begin{math}P_{s}^{\text {sup}}\end{math} and \begin{math} P_{s}^{\text {unsup}}\end{math} denote the predictions of the student model, while \begin{math}P_{t}^{\text {unsup}} \end{math} refers to the pseudo-labels generated by the teacher model. \begin{math}Y^{\text {sup }}\end{math} denotes the ground-truth annotations. Finally, the teacher model is gradually updated from the student model using the EMA, as shown below:
\begin{equation}
\begin{split}
\Theta ^{t}_{i} \leftarrow \alpha \Theta^{t}_{i-1}+(1-\alpha) \Theta^{s}_{i},
\end{split}
\end{equation}
where \begin{math} i \end{math} represents the training step, \begin{math} \Theta_{t} \end{math} and \begin{math} \Theta_{s} \end{math} are the parameters of the teacher and student models, respectively, and \begin{math} \alpha \end{math} is the momentum decay, set to 0.999. 

To further improve detection performance, we introduce additional guidance from the DINOv2\cite{oquab2023dinov2} in terms of both pseudo-label generation and feature extraction. Our main contribution consists of two novel components, VPM strategy and DVA method, which will be elaborated in the following sections.

\subsection{VFM as Free Guides}
\label{free}

Our VG-DETR leverages feature maps extracted from a frozen VFM to guide pseudo-label generation and improve the feature representation of the detector. Although the VFM is solely used for inferring feature maps, it still incurs considerable computational overhead due to the high input size of detection images (e.g., 800 pixels), which is significantly greater than that of classification tasks. Given that we have access to all target domain images and partial annotations, we move the entire inference pipeline of VFM to the offline stage, including feature extraction and referenced prototype generation (described in Subsection \ref{Mining}), and store the results on disk. During the training of the detector, these precomputed features are loaded directly, enabling a \emph{free lunch} style of guidance without additional runtime cost.

DINOv2\cite{Dosovitskiy_Beyer_Kolesnikov_Weissenborn_Zhai_Unterthiner_Dehghani_Minderer_Heigold_Gelly} is a VFM with strong generalization capability, trained on large-scale data using self-supervised learning. It has been show that the features extracted by DINOv2 are effective for remote sensing scenarios~\cite{bou2024exploring}. Moreover, it has been widely applied to various downstream visual tasks across different domains without fine-tuning, including remote sensing~\cite{zheng2024segment,fu2024cross} and medical imaging~\cite{baharoon2023evaluating,song2024dino}. Considering that disk storage is relatively inexpensive and the stored features (approximately 10 MB per image) are not required during deployment, we adopt DINOv2 ViT-L as our frozen encoder and extract target-domain feature maps, denoted as \begin{math} F^i \in \mathbb{R}^{H \times W \times d} \end{math}, where \begin{math} i \end{math} indexes the image, and \begin{math} H \end{math}, \begin{math} W \end{math}, and \begin{math} d \end{math} denote the height, width, and channel dimension of the feature map, respectively. We extract feature maps from the original (unaugmented) images. When spatial locations are aligned, these VFM feature maps provide effective guidance to the student model trained on strongly augmented inputs.

\subsection{VPM Strategy}
\label{Mining}

Maintaining high-quality and high-quantity pseudo-labels is crucial for performance improvement in the self-training paradigm. Some methods rely on manually set fixed thresholds, which require tedious hyperparameter tuning and are often suboptimal. Other approaches adopt dynamic thresholding, which inevitably excludes a considerable number of potentially correct predictions in semi-supervised tasks, as it typically maintains a relatively high threshold. To address this issue, we innovatively leverage a foundation model to identify likely correct samples from those discarded, thereby adaptively generating pseudo-labels with both high quality and quantity.

\subsubsection{Referenced Class-wise Prototype Extraction} Since a portion of labeled samples is available, we exploit the label information to extract class-wise object prototypes and background prototypes as references, which are then used to recover potentially correct pseudo-labels from low-confidence predictions. Specifically, given a feature map \begin{math} F^i\in\mathbb{R}^{H\times W \times d} \end{math} extracted as described in Subsection \ref{free}, we treat the labeled bounding boxes as object proposals. First, we use ROI Align\cite{he2017mask} to obtain indstance features \begin{math} F^i_{ins}\in\mathbb{R}^{N_{i}\times d} \end{math}, where \begin{math} N_{i} \end{math} denotes the number of labeled objects in the \begin{math} i \end{math}-th sample. Evidently, we have access to all labeled samples, allowing us to conveniently extract the instance features \begin{math} F_{ins}\in\mathbb{R}^{N\times d} \end{math} corresponding to all labeled samples. 
Then, considering that objects within the same category in remote sensing scenes often exhibit multimodal distributions, we apply the K-Means algorithm to all instance-level features of the same category, grouping them into \begin{math} K \end{math} clusters by minimizing the following error:
\begin{equation}
\sum_{k=1}^{K} \sum_{j=1}^{N_k} \| f_{j} - \mathbf{C}_k \|_2^2,
\end{equation}
where \begin{math} \| \cdot \|_2^2 \end{math} denotes the squared Euclidean distance, \begin{math} f_{j} \end{math} represents the \begin{math} j \end{math}-th instance feature vector from \begin{math} F^i_{ins} \end{math}, and \begin{math} \mathbf{C}_k \end{math} represents the \begin{math} k \end{math}-th centroid, which can be described as:
\begin{equation}
\mathbf{C}_k = \frac{1}{N_k} \sum_{i \in \mathcal{C}_k} f_{i},
\end{equation}
where \begin{math} N_{k} \end{math} denotes the number of feature vectors belonging to \begin{math} \mathbf{C}_k \end{math}. Note that the number of cluster \begin{math} K \end{math} is not the same as the number of categories, and the ablation study of the value of \begin{math} K \end{math} is explored in Section \ref{sec:number_components}. For background prototype extraction, we generally follow the same procedure described above. In this process, the background annotations are generated by simply flipping the ground-truth boxes and removing overlapping regions based on IoU, in order to simulate erroneous detections. Finally, for each category \begin{math} c \in \{1, 2, \dots, C,BG\} \end{math},where \( BG \) denotes the background class, we obtain a set of \begin{math} K \end{math} reference prototypes, denoted by \begin{math} P^c \in \mathbb{R}^{K \times d} \end{math}. The complete set of prototypes is denoted as
\begin{math} P_{ref} = \{P^1, P^2, \dots, P^C,P^{BG}\}\end{math}. As illustrated in Fig. \ref{fig_1} (a), the entire process is executed offline and imposes no additional computational burden during training.

\begin{figure}[tbp]
\centering
\includegraphics[width=\linewidth]{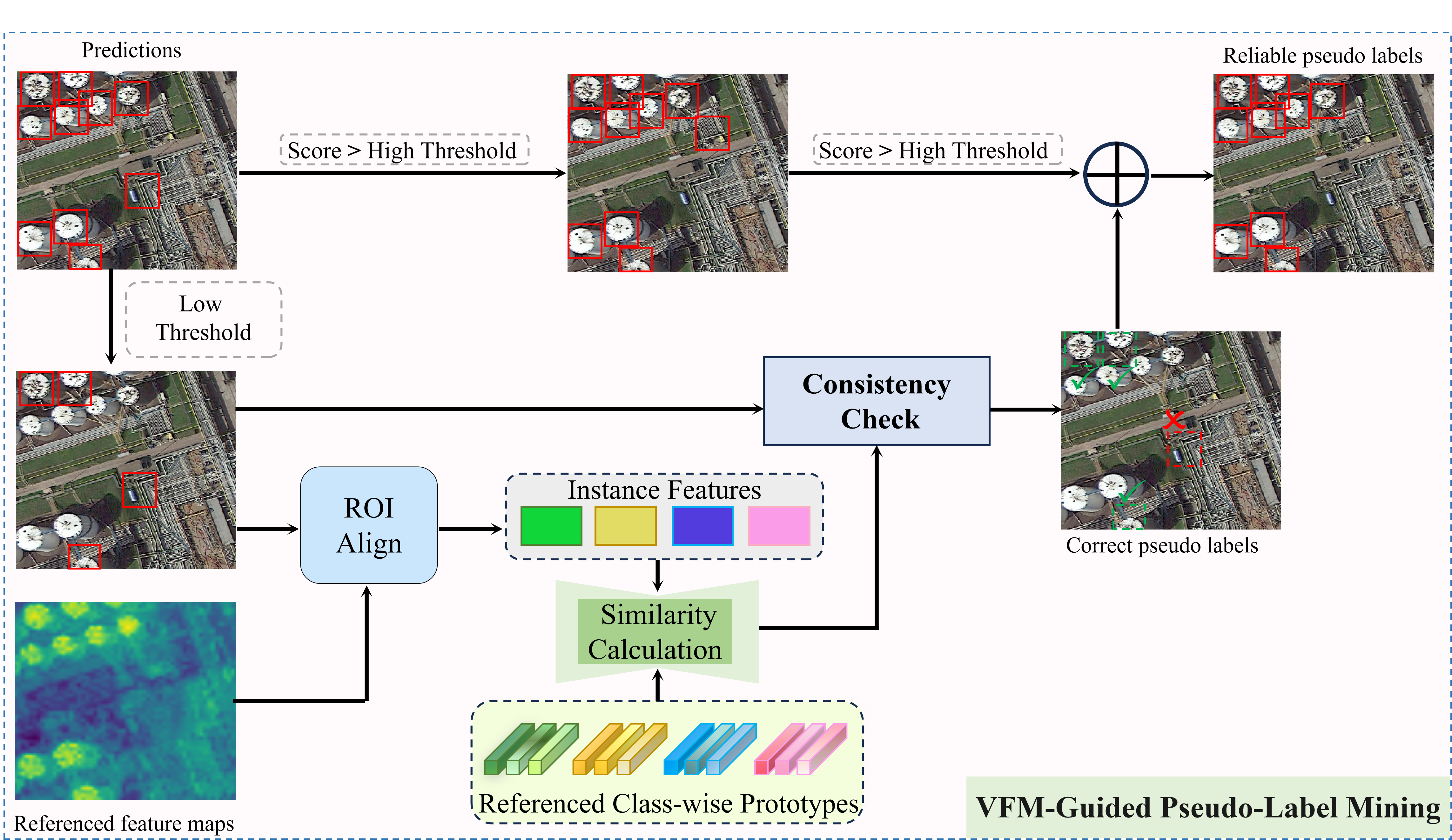}
\caption{ Details of the proposed VPM strategy. The strategy leverages reference prototypes extracted from the VFM to evaluate the reliability of the generated pseudo-labels and to identify potentially correct predictions within low-confidence outputs.}
\label{fig_2}
\end{figure}

\subsubsection{Pseudo-label Mining Strategy} For unlabeled target-domain samples, we leverage the extracted reference prototypes \begin{math} P_{ref} \end{math} to online mine potentially correct pseudo-labels with low confidence, as illustrated in Fig. \ref{fig_2}. Specifically, given an unlabeled sample and its corresponding feature map \begin{math} F^i \end{math} from the VFM, we also apply ROI Align to extract the instance features \begin{math} F^i_{ins} \end{math}, treating the predicted bounding boxes as object proposals. Then, we compute the cosine similarity between the \begin{math} F^i_{ins} \end{math} and the referenced prototypes \begin{math} P_{ref} \end{math}, as shown in the following equation:
\begin{equation}
\text{sim}(f_j) = \frac{f_j \cdot P_{ref}}{\|f_j\|_2 \|P_{ref}\|_2},
\end{equation}
where \begin{math} f_{j} \in F^i_{ins} \end{math} denotes the \begin{math} j \end{math}-th instance feature vector. Furthermore, we assess the reliability of the predicted results based on the computed cosine similarity. If the highest computed similarity of a feature vector exceeds a predefined threshold (empirically set to 0.5) and its corresponding category matches the predicted class from the detector, the prediction is considered reliable and included in the pseudo-labels; otherwise, it is discarded.

It is worth noting that using feature maps extracted from the VFM to assess pseudo-label reliability may be imprecise, particularly in complex remote sensing scenarios with limited labeled data. Therefore, our strategy focuses on mining pseudo-labels with relatively low confidence. Specifically, we adopt a dual-threshold scheme in which pseudo-labels are evaluated for reliability only when their confidence scores fall between a lower and an upper threshold. The upper threshold is dynamically determined based on our previous work\cite{10474037}, thereby eliminating the need for additional hyperparameter tuning. The optimal lower threshold is identified through ablation studies \cref{sec:vpm_low}, and we demonstrate that our strategy remains robust across a wide range of lower threshold values.

\subsection{DVA Module}

\begin{figure}[bp]
\centering
\includegraphics[width=\linewidth]{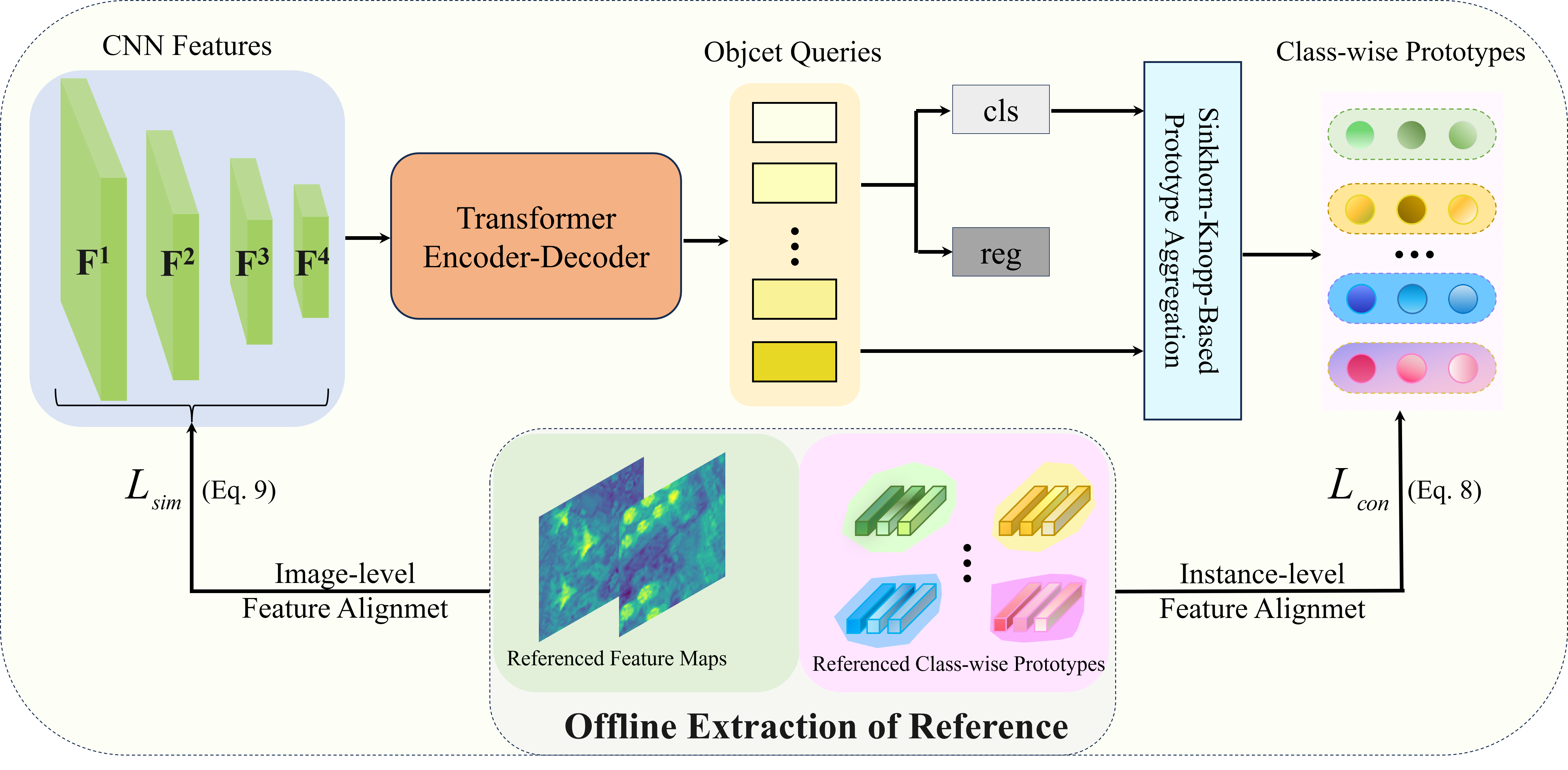}
\caption{Details of the proposed DVA module. The module leverages reference feature maps and class-wise prototypes extracted from the VFM to align detector representations with VFM embeddings at both the image and instance levels.}
\label{fig_3}
\end{figure}
It is evident that the VFM, pre-trained on large-scale data, exhibits stronger generalization capabilities than a detector trained solely on source-domain data. To this end, we leverage the feature maps from the VFM as alignment references to guide the detector in learning robust representations at both image and instance levels, thereby reducing the domain gap, as illustrated in Fig. \ref{fig_3}. Our alignment method imposes no structural constraints: it neither requires the VFM to extract multi-scale features aligned with the detector’s backbone, nor demands architectural consistency (e.g., requiring both models to adopt the same architecture, such as CNNs or ViTs).

\subsubsection{Instance-level Feature Alignment} As discussed in subsection \ref{Mining}, we obtain class-wise referenced prototypes \begin{math} P \end{math} by aggregating feature maps from the VFM. These prototypes are further utilized as references to guide object queries in capturing more generalizable representations. Following \cite{10841964}, we assign a class label to each object query using the detector’s predictions and then aggregate class-wise prototypes. Unlike prior work, we employ a Sinkhorn-based soft-clustering scheme that extracts multiple fine-grained prototypes per class. Consequently, our method aligns features to \begin{math} K \end{math} reference prototypes for each class, better capturing the multimodal object distributions typical of remote-sensing imagery.

Given the object query features \begin{math} Q\in\mathbb{R}^{B\times N \times d'} \end{math}, where \begin{math} B \end{math} is the batch size, \begin{math} N \end{math} is the number of object queries per image, and \begin{math} d' \end{math} is the channel dimension, the corresponding classification logits are represented as \begin{math} P_{logit}\in\mathbb{R}^{B\times N \times C} \end{math}, where \begin{math} C \end{math} denotes the number of categories. We first apply the sigmoid activation to \begin{math} P_{logit} \end{math} to obtain the confidence scores for object query. Then, the category with the highest score for each query is selected as the label, ultimately yielding the corresponding category matrix  \begin{math} \hat{Y} \in \mathbb{R}^{B \times N} \end{math}.

Based on this, we introduce a Sinkhorn-based soft clustering to model class-wise multiple prototypes of object queries. This mechanism is unsupervised, fully differentiable, enabling each query feature to be softly assigned to multiple cluster components. Specifically, for each class \begin{math} c \in C \end{math}, we first collect the set of query features predicted to belong to that class:
\begin{equation}
Q^{(c)} = \left\{ q_i \in Q \,\middle|\, \hat{y}_i = c \right\}.
\end{equation}
We then construct a soft assignment matrix \begin{math} A^{(c)}\in\mathbb{R}^{N_c \times K} \end{math}, where \begin{math} N_c = |Q^{(c)}| \end{math}, and \begin{math} K \end{math} is the number of prototype components assigned per class. The matrix is initialized by sampling from a standard normal distribution. To achieve a balanced assignment between object query features and prototype components, the sinkkhorn-knopp algorithm is applied to iteratively normalize \begin{math} A^{(c)} \end{math}. Finally, the \begin{math} k \end{math}-th prototype \begin{math} p_c^{k} \end{math} for class \begin{math} c \end{math} is computed as the weighted mean of the corresponding query features:
\begin{equation}
p_c^{k} = \frac{1}{\sum_{i=1}^{N_c} A^{(c)}_{i,k}} \sum_{i=1}^{N_c} A^{(c)}_{i,k} \cdot q^{(c)}_i.
\end{equation}

Based on the above computations, we obtain a set of class-wise prototypes \begin{math} P\in\mathbb{R}^{C\times K \times d} \end{math} during each training batch. We then perform contrastive learning between \begin{math} P \end{math} and \begin{math} P_{ref} \end{math}, treating sub-prototypes from corresponding positions of the same category as positive samples, and all others as negative samples. The contrastive loss is formulated as:
\begin{equation}
\begin{split}
\mathcal{L}_{\text{con}}
  &= -\frac{1}{CK} \sum_{i=1}^{C} \sum_{k=1}^{K}
     \log
     \frac{\exp\!\bigl(p_{i,k}\, MLP(p^{\text{ref}}_{i,k})\bigr)}
          {\displaystyle
           \sum_{j=1}^{C}\sum_{n=1}^{K}
           \exp\!\bigl(p_{i,k}\, MLP(p^{\text{ref}}_{j,n})\bigr)},
\end{split}
\end{equation}
where "\begin{math} \cdot \end{math}" is used to measure the similarity between queries from different domains, and \begin{math} MLP (\cdot) \end{math} represents a 3-layer perceptron with ReLU activation functions, used to align the channel dimension.

\subsubsection{Image-level Feature Alignment} We further use the feature maps \begin{math} F^i\in\mathbb{R}^{H\times W \times d} \end{math} from the VFM as references to align the detector’s backbone by computing their similarity. This brings two main benefits: a) Enhances instance-level feature alignment. Since the VFM typically adopts a ViT-based architecture that differs significantly from the CNN-based backbone used in most detectors, the features extracted from the two are inherently heterogeneous, making instance-level alignment suboptimal. Performing image-level feature alignment can increase the similarity between features from the VFM and the detector, thereby alleviating this gap. b) Improves the quality of backbone features. The VFM is capable of generating more fine-grained and robust features. By learning the feature generation patterns of the VFM, the detector can improve its representation of foreground objects and more effectively suppress interference from complex backgrounds.

In this paper, we intuitively leverage single-level feature maps from the VFM to guide the multi-level feature maps of the detector backbone, which yields promising results. This improvement can be attributed to our use of similarity-based alignment instead of adversarial training. Specifically, given a feature map \begin{math} F_s\in\mathbb{R}^{(H \times W \times d')} \end{math} from a layer of the student model, we first apply a \begin{math} 1 \times 1 \end{math} convolution to match the channel dimension \begin{math} d \end{math}. Next, bilinear sampling is employed to align the height \begin{math} H \end{math} and width \begin{math} W \end{math} of the VFM feature map with those of \begin{math} F_s \end{math}. Finally, the alignment loss is defined based on the similarity between the two normalized feature maps:
\begin{equation}
\begin{split}
\mathcal{L}_{\text{sim}}
  &= \frac{1}{HW} \sum_{h,w=1}^{HW}
     \Bigl(
       1 -
       \frac{
         \text{interp}(F^i)^{\top}\, \text{Conv}(F_s)
       }{
         \|\text{interp}(F^i)\|_2
         \;
         \|\text{Conv}(F_s)\|_2
       }
     \Bigr),
\end{split}
\end{equation}
where \begin{math} \text{interp}(\cdot) \end{math} denotes bilinear interpolation, and \begin{math} \text{Conv} \end{math} represents a \begin{math} 1 \times 1 \end{math} convolution.

\subsection{Network training} 

The proposed VG-DETR is a semi-supervised framework for source-free object detection in remote sensing imagery, which innovatively incorporates a VFM to provide additional guidance for generating reliable pseudo labels and enhancing the robustness of feature representation. The overall loss function combines the detection loss (Eq. 2), the contrastive learning loss (Eq. 8), and the similarity loss (Eq. 9) as follows:

\begin{equation}
\begin{aligned}
\mathcal{L} &=\mathcal{L}_{det} + \lambda_{con}\mathcal{L}_{con} + \lambda_{sim}\mathcal{L}_{sim},
\end{aligned}
\end{equation}
where \begin{math} \lambda_{con} \end{math} and \begin{math} \lambda_{sim} \end{math} are hyperparameters that balance the respective weights of the losses associated with the instance-level and image-level alignment terms.

\section{EXPERIMENTS}

\begin{table*}[t!]
  \centering
  \caption{Experimental results (\%) of the scenario between different optical satellites: xView $\rightarrow$ DOTA.}
  \renewcommand{\arraystretch}{1.0}
  \begin{tabular}{c|c|c|c|c c c|c}
    \specialrule{0.1em}{0pt}{0pt}  
    \multicolumn{8}{c}{\textbf{xView $\rightarrow$ DOTA1.0}} \\
    \specialrule{0.08em}{0pt}{0pt}
    \textbf{Setting} & \textbf{Method} & \textbf{Detector} & \textbf{Labeled} & \textbf{Plane} & \textbf{Storage-tank} & \textbf{Ship} & $mAP_{50}$ \\
    \specialrule{0.08em}{0pt}{0pt}
    \multirow{2}{*}{-}     & Source-DINO~\textcolor{gray}{\footnotesize [ICLR'23]}\cite{zhang2022dino} & \multirow{2}{*}{DINO} & \multirow{2}{*}{-}      & 63.5 & 33.2 & 57.0 & 51.2 \\
                           & Oracle-DINO~\textcolor{gray}{\footnotesize [ICLR'23]}\cite{zhang2022dino} &                        &                              & 94.8 & 77.8 & 74.3 & 82.3 \\
    \cmidrule{1-8}
    \multirow{4}{*}{UDAOD} & DAF~\textcolor{gray}{\footnotesize [CVPR'18]}\cite{Chen_Li_Sakaridis_Dai_Van_Gool_2018}   & Faster R-CNN         & \multirow{4}{*}{-}      & 57.3 & 36.4 & 34.3 & 42.7 \\
                           & EPM~\textcolor{gray}{\footnotesize [ECCV'20]}\cite{Hsu_Tsai_Lin_Yang_2020}   & FCOS                 &                          & 60.5 & 43.9 & 59.8 & 54.8 \\
                           & AQT~\textcolor{gray}{\footnotesize [IJCAI'22]}\cite{Huang_Lu_Lin_Xie_Lin}    & Deformable DETR      &                          & 68.2 & 46.8 & 59.9 & 58.3 \\
                           & RST~\textcolor{gray}{\footnotesize [TGRS'24]}\cite{10474037}                 & Deformable DETR      &                          & 75.7 & 54.7 & 59.7 & 63.3 \\
    \cmidrule{1-8}
    \multirow{4}{*}{SFOD}  & IRG~\textcolor{gray}{\footnotesize [CVPR'23]}\cite{vs2023instance}          & Faster R-CNN         & \multirow{4}{*}{-}      & 66.3 & 39.9 & 47.2 & 51.1 \\
                           & LPLD~\textcolor{gray}{\footnotesize [ECCV'24]}\cite{yoon2024enhancing}      & Faster R-CNN         &                          & 71.9 & 49.7 & 55.1 & 58.9 \\
                           & DRU~\textcolor{gray}{\footnotesize [ECCV'24]}\cite{khanh2024dynamic}        & Deformable DETR      &                          & 68.2 & 34.7 & 37.9 & 47.1 \\
                           & DINO + MT (Source-Free)                                                     & DINO                 &                          & 72.7 & 48.9 & 60.4 & 60.7 \\
    \specialrule{0.08em}{0pt}{0pt}
    \multirow{3}{*}{Fine-tuning} & \multirow{3}{*}{DINO~\textcolor{gray}{\footnotesize [ICLR'23]}\cite{zhang2022dino}} & \multirow{3}{*}{DINO} & 1\%   & 69.6 & 60.0 & 63.8 & 64.6 \\
                                 &                                                                       &                       & 5\%   & 81.9 & 63.9 & 65.8 & 70.5 \\
                                 &                                                                       &                       & 10\%  & 84.4 & 69.1 & 68.1 & 73.8 \\
    \cmidrule{1-8} 
    \multirow{16}{*}{SSOD} & \multirow{3}{*}{Pseco~\textcolor{gray}{\footnotesize [ECCV'22]}\cite{liu2022unbiased}} & \multirow{3}{*}{Faster R-CNN} & 1\%   & 69.8 & 58.2 & 59.6 & 62.5 \\
                           &                                                                               &                       & 5\%   & 79.1 & 66.3 & 66.7 & 70.7 \\
                           &                                                                               &                       & 10\%  & 80.1 & 68.5 & 68.3 & 72.3 \\
   \cmidrule{2-8}
                           & \multirow{3}{*}{DINO + MT (Semi-Supervised)}                              & \multirow{3}{*}{DINO} & 1\%   & 80.6 & 60.9 & 58.5 & 67.2 \\
                           &                                                                             &                       & 5\%   & 88.7 & 70.5 & 63.7 & 74.6 \\
                           &                                                                             &                       & 10\%  & 89.5 & 71.8 & 67.7 & 76.3 \\
    \cmidrule{2-8}
                           & \multirow{3}{*}{Semi-DETR~\textcolor{gray}{\footnotesize [CVPR'23]}\cite{zhang2023semi}} & \multirow{3}{*}{DINO} & 1\%   & 84.8 & 61.3 & 57.1 & 67.7 \\
                           &                                                                                           &                       & 5\%   & 88.9 & 68.9 & 67.5 & 75.1 \\
                           &                                                                                           &                       & 10\%  & 89.6 & 72.7 & 67.9 & 76.8 \\
      \cmidrule{2-8}
                           & \multirow{3}{*}{MCL~\textcolor{gray}{\footnotesize [AAAI'25]}\cite{wang2025multi}}         & \multirow{3}{*}{FCOS} & 1\%   & 78.0 & 56.8 & 66.4 & 67.1 \\
                           &                                                                                           &                       & 5\%   & 81.1 & 63.9 & 68.7 & 71.2 \\
                           &                                                                                           &                       & 10\%  & 84.0 & 68.6 & 69.3 & 73.9 \\
     \cmidrule{2-8}
        \noalign{\vskip -0.6ex}  

                           & \multirow{3}{*}{VG-DETR (Ours)}                                                           & \multirow{3}{*}{DINO} & \hl{1\%}   & \hl{83.3} & \hl{62.1} & \hl{64.9} & \hl{\textbf{70.5}} \\
                           &                                                                                           &                       & \hl{5\%}   &  \hl{90.1} & \hl{73.3} & \hl{68.6} & \hl{\textbf{77.5}} \\
                           &                                                                                           &                       & \hl{10\%}  & \hl{90.3} & \hl{74.5} & \hl{70.4} & \hl{\textbf{78.4}} \\
    \specialrule{0.1em}{0pt}{0pt}  
  \end{tabular}
  \label{tab:xview_dota}
\end{table*}

This section details our experimentation, which includes datasets and evaluation metric, implementation details, comparisons with state-of-the-art approaches, as well as ablation studies and analysis. Detailed discussions on each of these aspects are provided in the subsequent subsections.

\subsection{Datasets and Evaluation Metric}

In this work, we establish a benchmark for effectively evaluating the SFOD task in remote sensing scenarios, covering three distinct cross-domain settings utilizing seven publicly available datasets. Specifically, these include: (1) cross-satellite, from xView to DOTA; (2) synthetic-to-real, from GTA10K to DIOR; and (3) cross-modal, from HRRSD to SSDD. In all scenarios, our data split strictly follows the official train-test partition protocols. 

\subsubsection{xView\protect\cite{Lam_Kuzma_McGee_Dooley_Laielli_Klaric_Bulatov_McCord_2018}}The dataset originates from the WorldView-3 satellite, which offers a spatial resolution of up to 0.3 meters and encompasses a wide range of scenes, from densely populated urban environments to remote rural areas. Following existing cross-domain detection tasks, xView is commonly used as a source domain dataset paired with the DOTA dataset to evaluate a detector’s cross-satellite adaptation capability. In the experiments, the subcategories are merged into broader categories to align with DOTA’s category settings, and only images containing airplanes, storage tanks, and ships are used for training.
 
\subsubsection{DOTA1.0\cite{Xia_Bai_Ding_Zhu_Belongie_Luo_Datcu_Pelillo_Zhang_2018}}The dataset, primarily derived from Google Earth with resolutions ranging from 0.1 to 4.5 meters, is widely used for various remote sensing object detection tasks. In our experiments, it serves as the target domain and is paired with the xView dataset to construct a cross-satellite adaptation scenario. The officially defined training set is used for detector training, while the validation set is employed for evaluation. Under the semi-supervised setting, we randomly select 1\%, 5\%, and 10\% of the training data as labeled subsets. We focus on three object categories: airplane, storage tank, and ship. Both xView and DOTA images are cropped into 800 × 800 patches with an overlapping stride of 100 pixels.

\subsubsection{SRSD}To construct a synthetic-to-real adaptation scenario for evaluating the generalization capability of our method, we integrate two synthetic remote sensing datasets: RarePlanes\cite{Shermeyer_Hossler_Etten_Hogan_Lewis_Kim_2021} and GTAV10k\cite{10247619}, which contain airplane and vehicle targets, respectively. The resulting dataset is referred to as the Synthetic Remote Sensing Dataset (SRSD). To balance the number of images between the two datasets, we use the test split of RarePlanes and the entire GTAV10k dataset to form our training set. The final constructed dataset consists of 19,196 image patches, covering a total of 73,444 airplane and 56,193 vehicles. Each image is resized such that its shorter side is set to 1024 pixels.

\subsubsection{DIOR\cite{Li_Wan_Cheng_Meng_Han_2020}}This dataset is a large-scale benchmark for object detection in optical remote sensing imagery, developed by Northwestern Polytechnical University. In our experiments, it is paired with SRSD to construct a synthetic-to-real domain adaptation scenario. We select images containing the aircraft and vehicle categories from the official training and test splits. Each image has a fixed resolution of 800×800 pixels.

\subsubsection{HRRSD\cite{zhang2019hierarchical}}The High-Resolution Remote Sensing Detection (HRRSD) dataset was collected by the University of Chinese Academy of Sciences and comprises images sourced from Google Earth and Baidu Maps, with spatial resolutions ranging from 0.15 to 1.2 meters. In our experiments, 2,165 images containing a total of 3,975 ships were selected. This dataset is used as the source domain and all samples are utilized for training.

\subsubsection{SSDD\cite{Li_Qu_Shao_2017}}The SAR Ship Detection Dataset (SSDD) is a widely used public dataset designed for SAR-based ship detection tasks. It contains 1160 images and 2456 ship instances, covering a wide range of variations in sea conditions, spatial resolutions, and sensor types. Following prior work\cite{Zhang_Li_Dong_Pan_Shi_2022,10474037,11016953}, SSDD is commonly paired with the HRRSD dataset to form a cross-modality domain adaptation benchmark. In our experiments, training is conducted on the officially provided training split, and performance is evaluated on the test set. Additionally, the short edge of each image is resized to 800 pixels.

We adopt mean Average Precision (mAP) at an IoU threshold of 0.5 as our evaluation metric, which is widely used in various cross-domain detection tasks. 

\subsection{Implementation Details}

Our VG-DETR is developed based on the DINO baseline detector, integrating a semi-supervised framework along with two proposed VFM-guided methods. To ensure a comprehensive and fair comparison, we evaluate our approach against both CNN-based and Transformer-based domain adaptation methods, using ResNet-50 pre-trained on ImageNet~\cite{Li_Fei-Fei_2009} as the backbone whenever applicable. In our experiments, we set the batch size to 4 and train the model using the Adam optimizer~\cite{Kingma_Ba_2014} with an initial learning rate of $2 \times 10^{-4}$ and $\beta_{1}$ set to 0.9. The model is trained for 12 epochs, and the learning rate is reduced by a factor of 0.1 at the 11\textsuperscript{th} epoch. For all cross-domain scenarios, the weights for the contrastive loss \begin{math} \lambda_{con} \end{math} and the similarity loss \begin{math} \lambda_{sim} \end{math} in Eq.~(10) are set to 0.1 and 1.0, respectively. All experiments are conducted on a single NVIDIA A6000 GPU with 48~GB of memory.

\begin{table*}[htbp]
  \centering
  \caption{Experimental results (\%) for the synthetic-to-real adaptation scenario: SRSD $\rightarrow$ DIOR.}
  \renewcommand{\arraystretch}{1.0}
  \begin{tabular}{c|c|ccc|ccc|ccc}
    \specialrule{0.1em}{0pt}{0pt}
    \multicolumn{11}{c}{\textbf{SRSD $\rightarrow$ DIOR}}\\
    \specialrule{0.08em}{0pt}{0pt}
    \multirow{2}{*}{\textbf{Setting}} & \multirow{2}{*}{\textbf{Method}}
      & \multicolumn{3}{c|}{\textbf{1\%}}
      & \multicolumn{3}{c|}{\textbf{5\%}}
      & \multicolumn{3}{c}{\textbf{10\%}}\\
    \cmidrule(lr){3-11}
      &       & Plane & Vehicle & $mAP_{50}$
              & Plane & Vehicle & $mAP_{50}$
              & Plane & Vehicle & $mAP_{50}$\\
    \specialrule{0.08em}{0pt}{0pt}
    Fine-tuning & DINO
        & 81.1 & 32.0 & 57.4
        & 82.6 & 40.6 & 62.4
        & 83.3 & 43.8 & 64.4\\
    \cmidrule{1-11}
    \multirow{5}{*}{SSOD}
      & Pseco
        & 70.9 & 25.6 & 48.3
        & 72.1 & 36.1 & 54.1
        & 72.0 & 38.9 & 55.4\\
      & DINO + MT
        & 86.2 & 30.8 & 58.5
        & 85.3 & 41.5 & 63.9
        & 85.2 & 46.3 & 65.7\\
      & Semi-DETR
        & 83.7 & \textbf{36.1} & 59.9
        & 84.0 & \textbf{46.8} & 65.4
        & 86.6 & 47.4 & 67.0\\
      & MCL
        & 67.0 & 35.3 & 51.2
        & 76.9 & 41.3 & 59.1
        & 74.6 & 43.7 & 59.2\\
      & \textbf{VG-DETR}
        & \cellcolor{LightGray}\textbf{86.4}
        & \cellcolor{LightGray}35.1
        & \cellcolor{LightGray}\textbf{60.7}
        & \cellcolor{LightGray}\textbf{85.4}
        & \cellcolor{LightGray}45.8
        & \cellcolor{LightGray}\textbf{65.9}
        & \cellcolor{LightGray}\textbf{87.1}
        & \cellcolor{LightGray}\textbf{47.7}
        & \cellcolor{LightGray}\textbf{67.4}\\
    \specialrule{0.1em}{0pt}{0pt}
  \end{tabular}
  \label{tab:srsd_dior_detail}
\end{table*}

\subsection{Comparing with State-of-the-Arts Approaches}

To demonstrate the effectiveness and generalization capability of the proposed VG-DETR, we evaluate its performance across representative cross-domain adaptation scenarios. All source and target domain datasets used in our settings are publicly available, aiming to establish a benchmark for domain adaptive object detection in the field of remote sensing. Specifically, the evaluation involves three domain adaptation settings: (1) cross-satellite adaptation from xView to DOTA1.0, (2) synthetic-to-real adaptation from SRSD to DIOR, and (3) cross-modal adaptation from HRRSD to SSDD. In the first cross-domain scenario, we conduct a comprehensive comparison by evaluating methods under three learning paradigms: unsupervised domain adaptation, source-free adaptation, and semi-supervised learning. Since semi-supervised approaches benefit from partial annotations and yield higher accuracy, we focus on comparisons with semi-supervised baselines in the remaining two scenarios to better highlight the effectiveness of VG-DETR.

In our experiments, “Source-DINO” refers to a model trained solely on the source domain and directly evaluated on the target domain. “Oracle-DINO” denotes a model initialized with source-pretrained weights and then fully trained on the complete target dataset, serving as an upper-bound reference. “Fine-tuning” indicates a setting in which the source-pretrained model is further trained using only a small portion of labeled data from the target domain. All methods in the SSOD settings adopt models initialized with source-trained weights, ensuring alignment with the SFOD protocol.

\subsubsection{Cross-satellite Adaptation} This scenario involves three mainstream paradigms in domain-adaptive object detection: UDAOD, SFOD, and SSOD, with comparative results shown in Table \ref{tab:xview_dota}. “DINO + MT” indicates the integration of the mean teacher framework into vanilla DINO and can be implemented in two settings. Because UDA methods can leverage supervisory signals from source-domain data during training, they typically outperform SFOD methods, which lack access to source data. Although some SFOD methods alleviate domain shift to a certain extent, their overall detection performance remains limited, and the training process is often unstable. With the introduction of a small amount of labeled target-domain data, detection performance improves significantly. Semi-supervised methods further outperform fine-tuning approaches by effectively leveraging unlabeled data.
In the SSOD setting, VG-DETR consistently outperforms the baseline across all supervision levels, validating the effectiveness of our proposed method. Specifically, VG-DETR attains mAP$_{50}$ scores of 70.5\%, 77.5\%, and 78.4\% with 1\%, 5\%, and 10\% of labeled target-domain data, respectively, markedly surpassing existing methods in all settings. Notably, under the extremely low-supervision condition with only 1\% labeled data, VG-DETR exceeds Semi-DETR by 2.8\% mAP, benefiting from the effective utilization of robust semantic features extracted from the VFM.

\subsubsection{Synthetic-to-real Adaptation} Table \ref{tab:srsd_dior_detail} reveals that our VG-DETR consistently surpasses both the fine-tuning baseline and four state-of-the-art SSOD works under every supervision level across the synthetic-to-real adaptation benchmark. With only 1\% of the target-domain annotations, VG-DETR attains a mAP$_{50}$ of 60.7\%, outperforming fine-tuned DINO by 3.3 percentage points and surpassing the strongest existing method, Semi-DETR, by 0.8 points. These gains are primarily due to the robust semantic cues provided by the VFM, which remain reliable even under extremely low annotation settings. At the category level, VG-DETR sets a new state of the art for plane detection across all annotation ratios, and matches or exceeds the best vehicle results once a slightly larger portion of target-domain labels is provided.  It is noteworthy that the MCL method performs worse with 10\% labels than with 5\% labels for the plane category, presumably because the detector overfits during training and consequently forgets the knowledge learned from the source domain. In contrast, DETR-based methods exhibit stronger robustness to over-fitting and catastrophic forgetting.

\subsubsection{Cross-modal Adaptation} We evaluate cross-modal adaptation on the HRRSD → SSDD benchmark. Table \ref{tab:srsd_dior_map} reports the results: VG-DETR attains the highest mAP at every labeling percentage. These findings compellingly demonstrate the superior generalization ability of our approach.

\begin{table}[htp]
  \centering
    \caption{Experimental results (\%) for the cross-modal adaptation scenario: HRRSD $\rightarrow$ SSDD.}
  \renewcommand{\arraystretch}{1.2}
  \begin{tabular}{c|c|ccc}
    \specialrule{0.1em}{0pt}{0pt}
    \multicolumn{5}{c}{\textbf{HRRSD $\rightarrow$ SSDD}} \\
    \specialrule{0.08em}{0pt}{0pt}
    \multirow{2}{*}{\textbf{Setting}} & \multirow{2}{*}{\textbf{Method}} 
      & \multicolumn{3}{c}{$mAP_{50}$} \\
    \cmidrule{3-5}
      &       & \textbf{1\%} & \textbf{5\%} & \textbf{10\%} \\
    \specialrule{0.08em}{0pt}{0pt}
    Fine-tuning & DINO & 55.1      &68.8       &73.5       \\
    \cmidrule(lr){1-5}
    \multirow{5}{*}{SSOD} 
      & Pseco & 57.1      & 69.2      &76.9       \\
      & DINO + MT             & 58.2      &69.4       & 75.7      \\
      & Semi-DETR    &61.2       &69.6       &76.3       \\
      & MCL                &60.7       & 69.7      &  77.1     \\
      & \textbf{VG-DETR}             &\cellcolor{LightGray}\textbf{61.4}       & \cellcolor{LightGray}\textbf{70.6}     &\cellcolor{LightGray}\textbf{78.0}       \\
    \specialrule{0.1em}{0pt}{0pt}
  \end{tabular}
  \label{tab:srsd_dior_map}
\end{table}

\subsection{Ablation Studies}
\label{Ablation Studies}

In this section, we first conduct ablation studies on each proposed component by removing specific modules to effectively demonstrate their individual contributions. Next, we focus on the VPM strategy introduced in Section \ref{Mining}, demonstrating that it outperforms both fixed threshold and dynamic threshold approaches. Furthermore, given that predictions with extremely low confidence are predominantly incorrect, we analyze the effect of varying the lower bound of the confidence threshold and show that our approach remains effective across a wide range of low-threshold settings. Finally, we investigate the impact of the number of prototypes modeled for the same category in both the pseudo-label mining strategy and the instance-level feature alignment. All of the above experiments are conducted under cross-satellite adaptation settings with only 5\% of labeled data, and the results are reported on the DOTA1.0 validation set. Moreover, we investigate the effect of different vision foundation models on feature extraction across multiple scenarios to further demonstrate the effectiveness of our method.

\subsubsection{Effectiveness of Each Component} Table \ref{tab:Ablation study} summarizes the contribution of each component. Training the detector solely on the source domain yields 51.2\% mAP, underscoring the domain gap. Fine-tuning this model with 5 \% of labelled target data boosts performance to 70.5\% mAP. Adding the semi-supervised MT framework with a pseudo-label confidence threshold of 0.3 yields a further 4.1\% gain, which confirms the benefit of exploiting unlabeled target images. Replacing the pseudo-labels generated by the fixed-threshold method with our VPM strategy, which mines reliable low-confidence predictions, yields an additional 1.9-point improvement. We then evaluate the proposed VFA. Aligning instance-level features alone is ineffective because the prototypes extracted from VFM differ markedly from the detector’s object-query embeddings. Conversely, image-level alignment refines feature extraction and lifts performance to 75.2\% mAP. Combining instance- and image-level alignment reaches 75.6\% mAP, indicating moderate complementarity. Finally, coupling VPM with the full VFA configuration yields VG-DETR, pushing performance to 77.5\% mAP.

\begin{table}[t]
  \centering
  \caption{Ablation study of VG-DETR}
  \label{tab:ablation}
  \setlength{\tabcolsep}{6pt}        
  \renewcommand{\arraystretch}{1.0} 

  \begin{tabular}{@{}lccccc@{}}
    \toprule
    \multirow{2}{*}{Setting} & \multirow{2}{*}{MT} & \multirow{2}{*}{VPM} & \multicolumn{2}{c}{DVA} & \multirow{2}{*}{$mAP_{50}$} \\
    \cmidrule(lr){4-5}
                             &                      &                      & Instance & Image & \\ \midrule
    Source-only              &                      &                      &          &       & 51.2 \\
    Fine-tuning              &                      &                      &          &       & 70.5  \\ \midrule
    + MT                     & \checkmark           &                      &          &       & 74.6  \\
    + VPM                    & \checkmark           & \checkmark           &          &       & 76.5  \\
    + DVA-Instance           & \checkmark           &                      & \checkmark &     & 74.5  \\
    + DVA-Image              & \checkmark           &                      &          & \checkmark & 75.2  \\
    + DVA-Inst \& Img        & \checkmark           &                      & \checkmark & \checkmark & 75.6  \\
    VG-DETR               & \checkmark           & \checkmark           & \checkmark & \checkmark & \textbf{77.5}  \\ 
    \bottomrule
  \end{tabular}
  \label{tab:Ablation study}%
\end{table}

\subsubsection{The Effect of the Pseudo-label Threshold} Selecting an appropriate threshold is essential for filtering out incorrect predictions during pseudo-label generation in the MT framework. As shown in Table \ref{tab:TV}, we first evaluate fixed-threshold methods by varying the threshold from 0.2 to 0.5. The results indicate that threshold choice is critical. Setting the confidence threshold to 0.4 yields the best performance 75.6\% mAP, an improvement of 5.0\% mAP over the 0.2 threshold. However, identifying an optimal threshold for every dataset is computationally expensive and often impractical. Dynamic thresholding techniques, although successful in fully unsupervised tasks, perform sub-optimally in semi-supervised settings. This is because the threshold is updated according to prediction confidence, which keeps it high and inevitably filters out many potentially correct predictions. Our proposed VPM strategy overcomes this limitation by using class-wise prototypes extracted from the VFM to assess the reliability of low-confidence pseudo-labels, thereby preserving both their quality and quantity. Consequently, our method attains a superior 76.5\% mAP.

\begin{table}[htbp]
  \centering
  \caption{Ablation study of threshold value}
    \begin{tabular}{crr}
    \toprule
    Method & \multicolumn{1}{c}{Threshold Value} & \multicolumn{1}{c}{$mAP_{50}$} \\
    \midrule
    \multirow{4}[1]{*}{Fixed} & \multicolumn{1}{c}{0.2} &70.6  \\
          & \multicolumn{1}{c}{0.3} &74.3  \\
          & \multicolumn{1}{c}{0.4} &75.6  \\
          & \multicolumn{1}{c}{0.5} &70.3  \\
    \midrule
    Dynamic\cite{10474037} & \multicolumn{1}{c}{-}      &74.8  \\
    VPM strategy &\multicolumn{1}{c}{-}      &\textbf{76.5}  \\
    \bottomrule
    \end{tabular}%
  \label{tab:TV}%
\end{table}%

\begin{table}[htbp]
  \centering
  \caption{Detection performance under different low-threshold values.}
  \renewcommand{\arraystretch}{1.2}
  \begin{tabular}{c|cccc}
    \toprule
    Method & 0.1 & 0.2 & 0.3 & 0.4 \\
    \midrule
    \textit{w/o} VPM strategy &  67.6  &  71.3  &  75.6  &  76.7  \\
    VG-DETR         &  69.7  &  74.1  &  \textbf{77.5}  &  76.9  \\
    \bottomrule
  \end{tabular}
  \label{tab:vpm_threshold}
\end{table}

\begin{figure}[htbp]
\centering
\includegraphics[width=\linewidth]{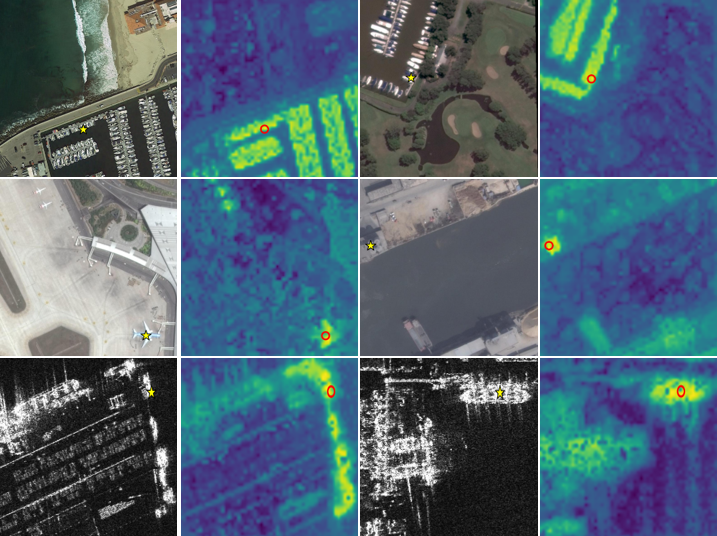}
\caption{The similarity between the reference regions and all other spatial locations in the feature maps extracted by DINOv2. The reference regions are marked with yellow stars, while the corresponding responses in the feature maps are highlighted with red circles.}
\label{fig_F_V}
\end{figure}

\begin{figure}[htbp]
\centering
\includegraphics[width=\linewidth]{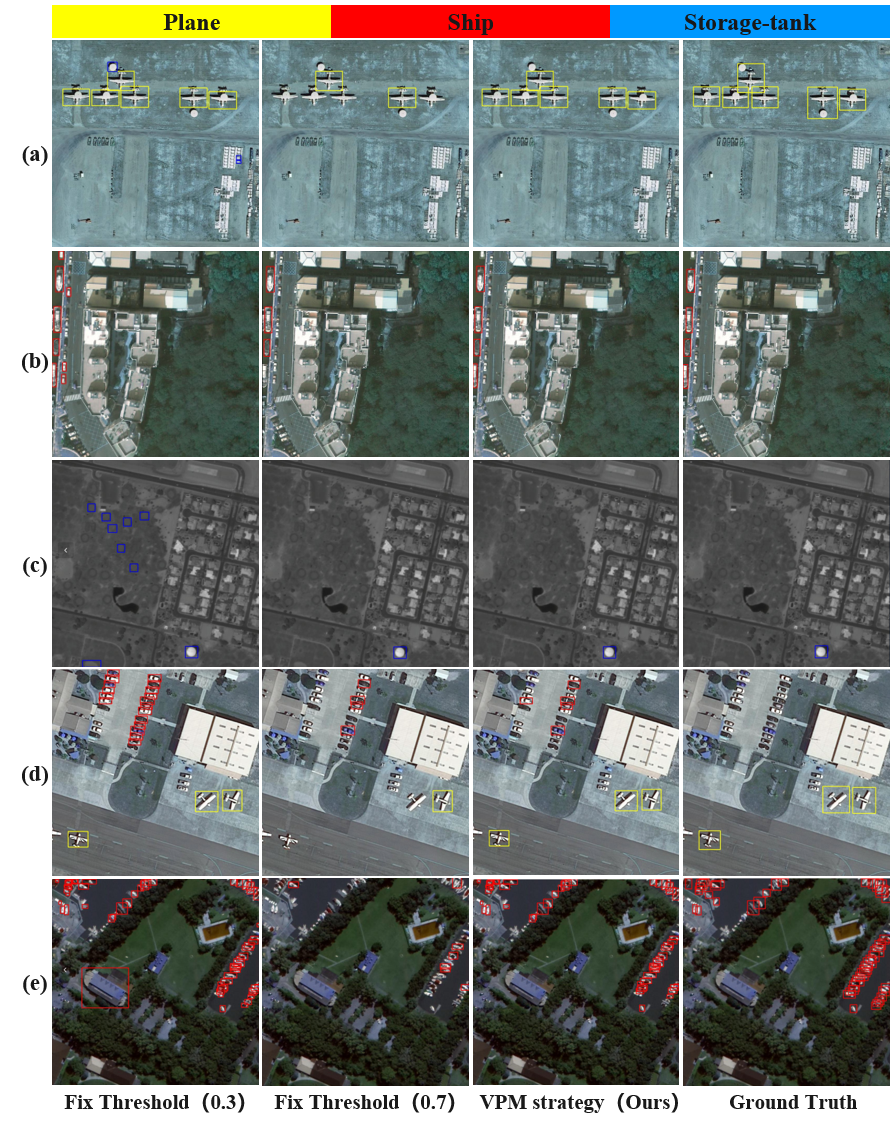}
\caption{Visualization of the pseudo-labels produced by the fixed-threshold method and by our VPM strategy, alongside the ground-truth annotations.}
\label{fig_PL_V}
\end{figure}

\label{Training_Stability}
\begin{figure}[htbp]
\centering
\includegraphics[width=\linewidth]{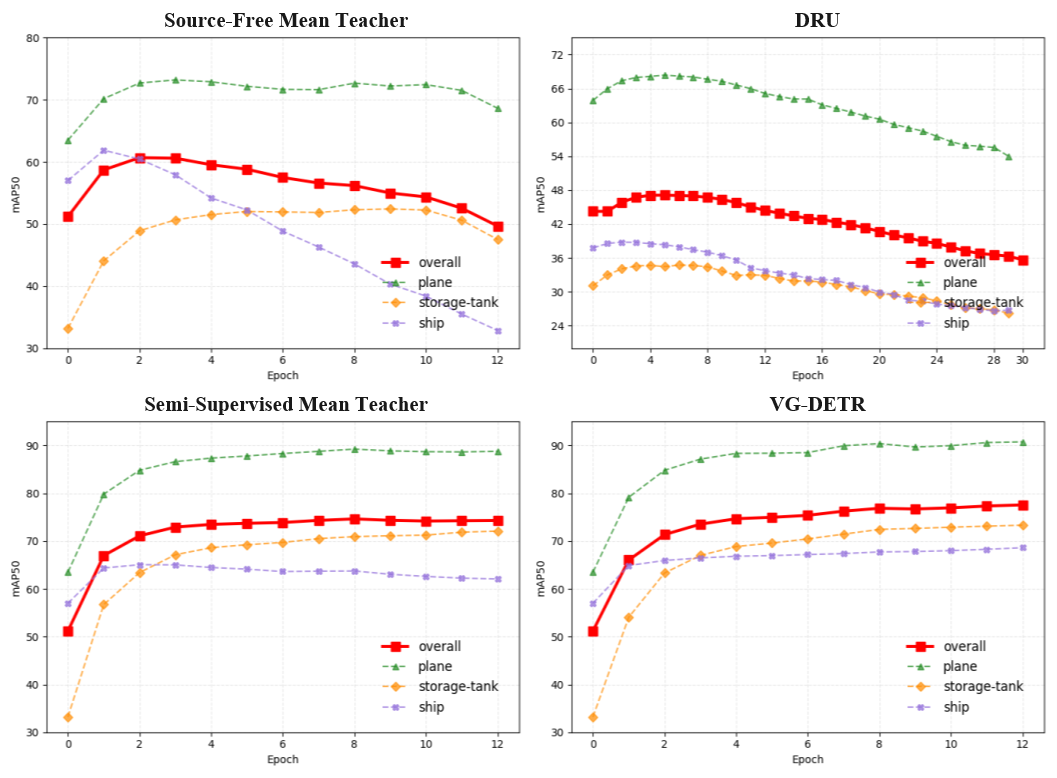}
\caption{Training curves of the four paradigms within the mean-teacher framework on the xView → DOTA task. mAP is tracked across epochs.}
\label{fig_stability}
\end{figure}

\begin{figure*}[htbp]
\centering
\includegraphics[width=\linewidth]{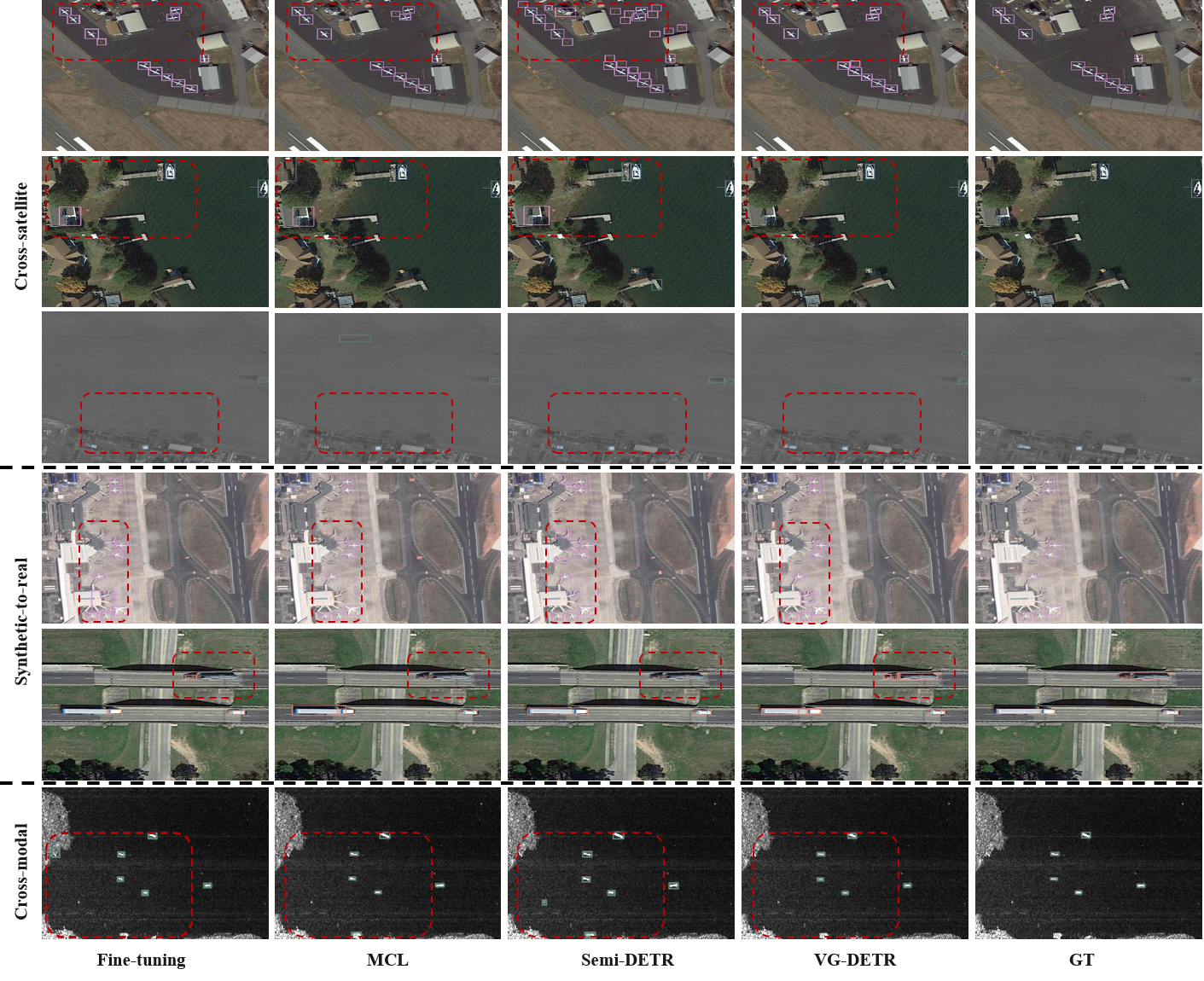}
\caption{Visual comparison across all cross-domain experimental scenarios, with the visualization threshold set to 0.2. “GT” denotes the ground truth.}
\label{fig_show_result}
\end{figure*}

\subsubsection{Robustness Under Low-Threshold Settings}\label{sec:vpm_low}We evaluate the proposed VPM strategy across a range of lower-bound confidence thresholds, as shown in Table \ref{tab:vpm_threshold}. Without VPM, increasing the threshold from 0.1 to 0.4 raises mAP from 67.6\% to 76.7\%, implying that most low-confidence predictions are false positives that hinder training. With VPM, VG-DETR exceeds the baseline at every threshold, indicating that the strategy effectively recovers true positives from low-confidence predictions and thus boosts detection accuracy. At a threshold of 0.4, performance drops slightly because the stricter criterion reduces the pool of exploitable low-confidence samples. The best result 77.5\% mAP is obtained at 0.3, showing that VG-DETR not only raises the upper bound of detection accuracy but also remains robust to threshold variations.

\subsubsection{The Effect of the Number of Components per Class Prototype}\label{sec:number_components} To investigate the impact of the number of components assigned to each class prototype, we conduct an ablation study by varying the number of clusters from 1 to 5. As shown in Table \ref{tab:number components}, using only a single component results in suboptimal performance, suggesting that one component is insufficient to capture the intra-class diversity. As the number of components increases, detection performance improves, reaching its peak when four Gaussian components are used per class, achieving a mAP of 77.5\%. This indicates that modeling each class prototype with multiple components allows the detector to better characterize fine-grained intra-class variations, thereby enhancing its feature representation and cross-domain robustness. However, further increasing the number of components beyond four leads to performance degradation, likely due to model overfitting or increased structural complexity, which in turn wpeakens its discriminative capability.

\begin{table}[tbp]
  \centering
  \caption{Ablation study of the number of components per class prototype.}
    \begin{tabular}{c|ccccc}
    \toprule
    Number & \multicolumn{1}{c}{1} & \multicolumn{1}{c}{2} & \multicolumn{1}{c}{3} & \multicolumn{1}{c}{4} & \multicolumn{1}{c}{5} \\
    \midrule
    $mAP_{50}$ &76.8 & 77.3 & 77.1 & \textbf{77.5} & 77.0 \\
    \bottomrule
    \end{tabular}%
  \label{tab:number components}%
\end{table}

\subsubsection{The Effect of Different Vision Foundation Models on Feature Extraction} We further evaluate the performance of our method with different VFMs across three cross-domain scenarios, as reported in Table \ref{tab:VFMs}. DINOv3~\cite{simeoni2025dinov3} outperforms DINOv2 in optical imagery, owing to its ability to capture finer-grained feature representations that are particularly beneficial for dense prediction tasks. The performance on SAR imagery remains largely comparable between the two models, primarily because DINOv3 does not produce finer feature representations than DINOv2, as neither model has been trained on SAR data. Under all experimental settings, our method consistently improves detection performance with different VFMs, thereby demonstrating its strong generalization ability.

\begin{table}[t]
  \centering
  \caption{Detection performance with different VFMs across multiple scenarios.}
  \setlength{\tabcolsep}{3pt}       
  \renewcommand{\arraystretch}{0.95}
  \footnotesize                     
  \begin{tabular*}{\columnwidth}{@{\extracolsep{\fill}} l|ccc|ccc|ccc}
    \toprule
    \multirow{2}{*}{VFM} & \multicolumn{3}{c|}{xView$\rightarrow$DOTA} & \multicolumn{3}{c|}{SRSD$\rightarrow$DIOR} & \multicolumn{3}{c}{HRRSD$\rightarrow$SSDD} \\
    \cmidrule(lr){2-4}\cmidrule(lr){5-7}\cmidrule(l){8-10}
     & 1\% & 5\% & 10\% & 1\% & 5\% & 10\% & 1\% & 5\% & 10\% \\
    \midrule
    N/A              & 67.2    &74.6     &76.3      & 58.5    & 63.9    & 65.7     & 58.2    &  69.4   &75.7      \\
    DINOv2-ViT-L/14  &70.5     & 77.5    &78.4      &\textbf{60.7}     &65.9   &67.4      &\textbf{61.4}     &70.6     & \textbf{78.0}     \\
    DINOv3-ViT-L/16  & \textbf{70.7}    & \textbf{77.6}   & \textbf{78.9}     &60.4     &\textbf{66.1}     &\textbf{67.6}      & 61.3    & \textbf{70.7}     &77.8     \\
    \bottomrule
  \end{tabular*}
  \label{tab:VFMs}
\end{table}

\subsection{Visualization and Analysis}

\subsubsection{Instance Feature Visualization} We evaluate the similarity between the reference regions (marked with yellow stars) and all other spatial locations in the feature maps extracted by DINOv2, as shown in Fig. \ref{fig_F_V}. It is evident that instances belonging to the same category exhibit substantially stronger responses than background regions, and this trend remains stable even in cluttered and dense scenes. Specifically, in optical imagery, the similarity maps capture fine-grained semantic structures despite complex texture interference. For SAR imagery, although the extracted features are of lower quality due to the absence of SAR-specific training, the model still produces semantically relevant responses while effectively suppressing background noise. These results demonstrate that our VG-DETR benefits from the guidance of vision foundation model, enabling it to reliably filter out unstable pseudo-labels and to enhance the robustness of object detection under cross-domain scenarios.

\subsubsection{Pseudo-label Visualization} Fig.~\ref{fig_PL_V} visualizes the pseudo-labels generated under fixed-threshold settings in the cross-satellite scenario xView~$\rightarrow$~DOTA, highlighting the effectiveness of the proposed VPM strategy. Following prior work~\cite{10474037,liu2024source}, the low and high confidence thresholds are typically set to 0.3 and 0.7, respectively. Due to the presence of domain gaps, detectors often produce low-confidence predictions. As a result, a threshold of 0.3 introduces numerous false positives, while a threshold of 0.7 discards many correct predictions. Our VPM strategy explores the ``grey zone'' between these two fixed thresholds, effectively filtering out false positives while preserving a large number of high-quality pseudo-labels, as illustrated in Fig.~\ref{fig_PL_V} (a)--(c). When handling scenarios that involve extremely small objects or densely packed regions, our method may produce evaluation inaccuracies, leading to residual noise in the generated pseudo-labels due to the limited spatial resolution of the feature maps, as shown in Fig.~\ref{fig_PL_V} (d)--(e). Nevertheless, the overall quality of the pseudo-labels remains superior to that obtained with fixed-threshold methods, which demonstrates the effectiveness of our approach.

\subsubsection{Training Stability} 

Fig.~\ref{fig_stability} summarizes the optimization trends of the four training paradigms on the xView → DOTA scenario, illustrating the epoch-wise evolution of mAP. In the first row, under the source-free training paradigm, the MT baseline attains its peak performance during the early epochs and is followed by a training collapse. Although DRU has been shown to alleviate this issue on benchmarks of natural scenes\cite{khanh2024dynamic}, it remains unavoidable in remote-sensing imagery, where complex backgrounds and densely packed objects exacerbate pseudo-label noise. 
In the second row, the semi-supervised MT curve converges stably without any sign of collapse, even when trained with only 5\% annotated data. This evidence indicates that introducing a small amount of labeled data not only corrects pseudo-label errors but also significantly enhances overall detection performance. Our VG-DETR achieves additional performance gains by integrating a VFM that further guides detector training.

\subsubsection{Detection Results} Fig.~\ref{fig_show_result} presents the visual results of VG-DETR across all evaluated domain-adaptation scenarios, alongside the finetuning baseline and several semi-supervised methods. In the cross-satellite adaptation task, our method achieves higher recall for the storage-tank class and other small objects while also reducing false positives caused by complex backgrounds. In the synthetic-to-real scenario, VG-DETR likewise provides more accurate detections, thereby lowering the false-positive rate and identifying challenging objects that the baseline fails to detect. For cross-modal adaptation, where a small amount of labeled data guides training, all detectors perform reasonably well. Our VG-DETR further suppresses false positives. The visual results correspond with the numerical evaluations, demonstrating VG-DETR’s superior performance and generalization across the domain-adaptation scenarios widely adopted in remote sensing.

\section{Conclusion}

We propose a Vision foundation model-Guided DEtection TRansformer (VG-DETR) for source-free object detection in remote sensing scenarios. Because source-free training can collapse, we adopt a semi-supervised framework that introduces a limited subset of labeled data, thereby preventing collapse while still achieving superior detection performance. To further enhance performance and generalization, we integrate a Vision Foundation Model (VFM) as a \emph{free-lunch} auxiliary supervisor, incurring almost no additional training cost. First, we devise a VFM-guided Pseudo-label Mining (VPM) strategy that leverages the VFM’s semantic priors to re-evaluate generated pseudo-labels. By recovering latent correct predictions from low-confidence outputs, the strategy markedly improves both the quality and quantity of pseudo-labels. In addition, we introduce a Dual-level VFM-guided Alignment (DVA) method that aligns detector features with VFM embeddings at the instance and image levels, thereby enhancing the robustness of the detector’s representations to domain gaps. Extensive experiments on diverse remote-sensing adaptation benchmarks demonstrate the outstanding performance of VG-DETR. In future work, we will explore domain-generalization and test-time adaptation techniques for object detection.

\bibliographystyle{IEEEtran}
\bibliography{References.bib}

\vfill

\end{document}